\documentclass[runningheads]{llncs}

\usepackage{eccv}

\usepackage{eccvabbrv}

\usepackage{graphicx}
\usepackage{booktabs}

\usepackage[accsupp]{axessibility}  

\usepackage{hyperref}

\usepackage{orcidlink}
\colorlet{shadecolor}{gray!90}
\usepackage[dvipsnames]{xcolor}
\usepackage{mathrsfs}
\usepackage{multirow}

\newcommand{\TODO}[1]{\textbf{\color{red}[TODO: #1]}}

\usepackage{comment}

\begin{document}

\title{RAP: Retrieval-Augmented Planner for Adaptive Procedure Planning in Instructional Videos} 

\titlerunning{RAP}

\author{Ali Zare \and
Yulei Niu\thanks{Corresponding author}\and
Hammad Ayyubi\and
Shih-Fu Chang}

\authorrunning{A.~Zare et al.}

\institute{Columbia University, New York, NY 10027, USA \\ \email{\{az2584,yn2338,ha2578,sc250\}@columbia.edu},
}

\maketitle

\begin{abstract}
  Procedure Planning in instructional videos entails generating a sequence of action steps based on visual observations of the initial and target states. Despite the rapid progress in this task, there remain several critical challenges to be solved: (1) Adaptive procedures: Prior works hold an unrealistic assumption that the number of action steps is known and fixed, leading to non-generalizable models in real-world scenarios where the sequence length varies.  (2) Temporal relation: Understanding the step temporal relation knowledge is essential in producing reasonable and executable plans. (3) Annotation cost: Annotating instructional videos with step-level labels (\ie, timestamp) or sequence-level labels (\ie, action category) is demanding and labor-intensive, limiting its generalizability to large-scale datasets. 
In this work, we propose a new and practical setting, called adaptive procedure planning in instructional videos, where the procedure length is not fixed or pre-determined. To address these challenges, we introduce Retrieval-Augmented Planner (RAP) model.
Specifically, for adaptive procedures, RAP adaptively determines the conclusion of actions using an auto-regressive model architecture. 
For temporal relation, RAP establishes an external memory module to explicitly retrieve the most relevant state-action pairs from the training videos and revises the generated procedures. 
To tackle high annotation cost, RAP utilizes a weakly-supervised learning manner to expand the training dataset to other task-relevant, unannotated videos by generating pseudo labels for action steps. 
Experiments on CrossTask and COIN benchmarks show the superiority of RAP over traditional fixed-length models, establishing it as a strong baseline solution for adaptive procedure planning.
\end{abstract}

\section{Introduction}
\label{sec:intro}
\begin{figure}{}
    \centering
    \includegraphics[width=1\linewidth]{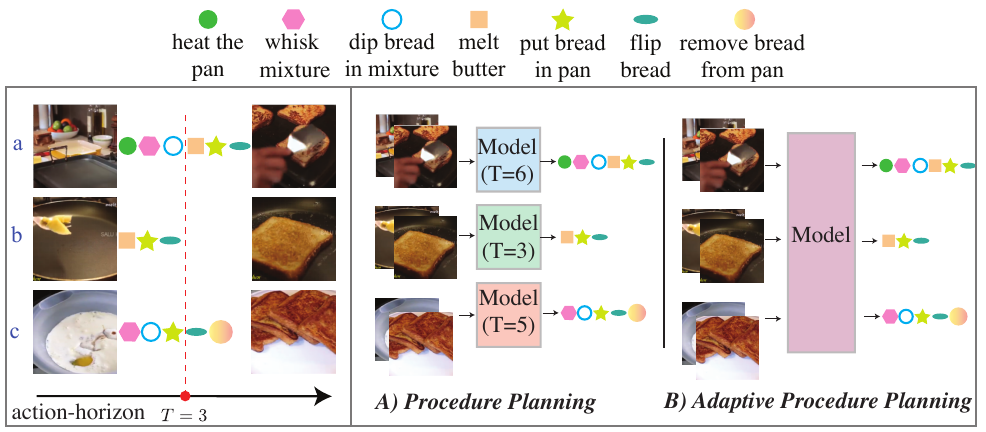}
    \caption{Examples of adaptive procedures in instructional videos. Procedure length varies with differences between initial and goal visual states, even for the same task. Previous works assume a fixed, known sequence length, requiring a separate model for each action-horizon. This figure contrasts traditional procedure planning (A) with our adaptive approach (B), which trains a unified model without knowing the procedure length in advance.}
    \label{fig:pp_fig}
\end{figure}

Procedure planning in instructional videos aims to generate an executive series of actions for completing tasks, transitioning from initial visual state to final visual state \cite{chang2020, bi2021, sun2022plate, p3iv, wang2023event, wang2023pdpp, li2023skip, niu2024schema}. While humans excel at such tasks, machine learning agents face challenges in achieving similar proficiency. Success in procedure planning requires an agent's deep understanding of actions, procedure states, involved objects, and their temporal relationships. Narrowing this performance gap is key to advancing autonomous systems, equipping them to adeptly perform complex, instruction-based tasks like cooking or repair.

Although recent works have made tremendous strides, there remain several challenges to real-world applications. First, previous works assume that the number of action steps is known and fixed. As depicted in \cref{fig:pp_fig}, the visual goal may be achieved through a small number of steps or a long procedure, depending on the differences between the initial and goal states. For example, it takes three steps to transform a visual of adding butter to a visual of flipped french toast in example b (scenario b implies pre-prepared mixture and pre-dipped bread, but visually requires actions like ``melt butter,'' ``put bread in pan,'' and ``flip bread.''), but one needs six and five steps to finish the cooking process based on the visuals of the mixture in scenarios a and c respectively. Previous models, trained on the unrealistic assumption of fixed prediction horizon, cannot figure out the complexity of state understanding and are hard to generalize to real-world applications. 
Secondly, there is a high correlation between consecutive action steps. For example, ``add'' is often followed by ``mix'', and ``wash'' is often followed by ``rinse''. Existing methods model this relationship implicitly via intermediate states, leading to sub-par results. 
Third, prior methods focused on fully annotated data~\cite{bi2021, chang2020, sun2022plate} -- action steps and intermediate visual states -- for supervision. However, this approach poses challenges in scaling up the training data due to its costliness.
Following works~\cite{p3iv, wang2023event,li2023skip, wang2023pdpp,niu2024schema} 
proposed to leverage text representations of action steps as language supervision,
which still necessitate sequence-level annotations, a time-consuming and expensive process involving watching entire instructional videos. 

To tackle these challenges, we introduce the setting of adaptive procedure planning in instructional videos and propose the Retrieval-Augmented Planner (RAP) model for this problem. Overall, RAP is designed to (1) adaptively predict procedures with variable lengths according to the complexity of visual states, (2) explicitly exploit step temporal relation via knowledge retrieval from the training data, and (3) learn procedural knowledge from additional unannotated videos. Specifically, RAP consists of an auto-regressive model as the base planner for adaptive procedure prediction, and a retrieval-augmented memory module to refine the generated procedures. The architecture is supplemented by weakly-supervised learning method that leverages both annotated video-procedure pairs and unannotated videos.

In addition, existing metrics for procedure planning, such as accuracy, fail to fairly evaluate variable length procedures (as described in \cref{sec:metrics}).  We address this by introducing a new metric edit-score. 
We evaluate our model on two widely used datasets for instructional videos, CrossTask \cite{crosstask} and COIN \cite{coin}. Our model demonstrates superior performance in the conventional setting and establishes a strong baseline in the new adaptive procedure planning setting.

Our main contributions are:
\vspace{-3mm}
\begin{itemize}
    \item We argue the unrealistic assumption of fixed-length plans in previous procedure planning works, and introduce a new setting adaptive procedure planning in instructional videos.
    \item We propose Retrieval-Augmented Planner that can generate adaptive procedures of variable lengths and retrieve step temporal relation knowledge from the training data.
    \item We introduce a weakly-supervised training approach to learn procedural knowledge from unannotated task-relevant videos.
\end{itemize}

\section{Related work}
\label{sec:related_work}

\subsection{Procedure Planning}
Procedure planning from instructional videos, involves generating effective plans for task completion based on start and final visual observations. Earlier methods utilized a two-branch architecture, alternating between action and state prediction, with models like recurrent neural networks  \cite{chang2020}, transformers \cite{sun2022plate}, and adversarial networks \cite{bi2021}. Recent approaches, such as P3IV's \cite{p3iv} memory-augmented Transformer and E3P, a prompting-based model \cite{wang2023event}, adopt one-branch transformer models. However, these methods are limited to predicting procedures of fixed lengths, lacking adaptability to tasks with varying action sequence length requirements. Our work introduces RAP, a novel one-branch approach designed to predict plans with variable action-horizons, addressing this limitation in the literature.

\subsection{Sequence Generation with Auto-regressive Transformers} 

Recent advancements in sequence generation using Transformer-based models \cite{vaswani2017attention, radford2019language} have established remarkable results in tasks involving sequence modeling, \eg natural language modeling \cite{radford2019language}. Procedure planning in its core, is a goal-conditioned sequence generation problem, thereby directly benefiting from the state-of-the-art sequence generation models. Leveraging these developments, our work incorporates an auto-regressive Transformer \cite{radford2019language} into the RAP architecture.

\subsection{Retrieval-Augmented Methods}

Augmenting language models (LM) with external memory \cite{lei2020mart, khandelwal2019generalization, grave2017unbounded, borgeaud2022improving, zhong2022training} has proved to significantly improve language sequence generation. P3IV \cite{p3iv} uses a memory component to enhance long-range sequence modeling in procedure planning. Similarly, we adopt a retrieval-based mechanism, drawing inspiration from Khandelwal \etal~\cite{khandelwal2019generalization}'s augmentation of a neural language model with a k-nearest neighbors (kNN) retrieval component, in our auto-regressive model to effectively leverage temporal prior information for predicting action sequences.

\subsection{Video Grounding}

Procedure planning suffers from a lack of annotated data. Earlier works heavily relied on fully annotated data (action-state timestamps) \cite{bi2021, chang2020, sun2022plate}. Recent approaches \cite{p3iv, wang2023event} have used language supervision during training which relies on less-costly, yet still resource-intensive language-level annotations. In this work, we introduce a weak language supervision framework that employs a video-language grounding algorithm to create pseudo-annotations from unannotated data. This approach is inspired by Dvornik \etal~\cite{dvornik2021drop} in sequence-to-sequence alignment research, utilizing a grounding algorithm that aligns video frame and action step representations.

\section{Technical approach}
\label{sec:Technical_approach}

In this section, we introduce our proposed framework Retrieval-Augmented Planner for adaptive procedure planning in instructional videos, and expand on a weakly-supervised learning method to leverage unannotated but task-relevant videos.
\subsection{Setting: Adaptive Procedure Planning}

In conventional formulation of procedure planning from instructional videos, a model is required to generate a plan $p =\{a_1, a_2,..., a_T\}$, given visual observations of the start $o_s$, and the goal $o_g$ states,  where $T$ is the plan's length (\ie, action-horizon).
Previous methods assume a fixed, known action-horizon, which is unrealistic for real-world applicability and limits the generalizability of planning models. As shown in \cref{fig:pp_fig}, we consider a more general and practical setting, \textit{adaptive procedure planning in instructional videos}, where the procedure length is not fixed.

\subsection{Model: Retrieval-Augmented Planner}
\label{sec:retrieval}

\begin{figure*}[t]
    \centering
    \includegraphics[width=0.9\linewidth]{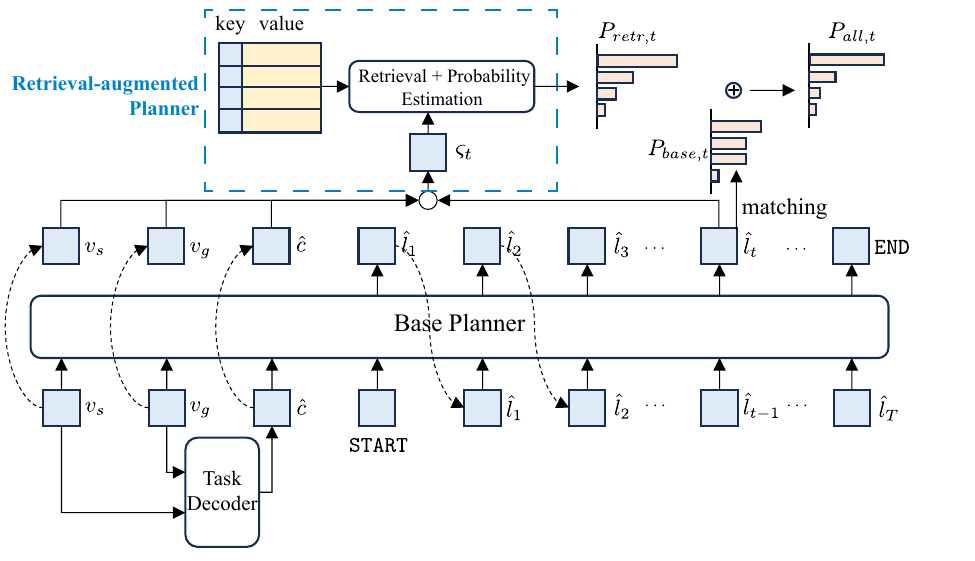}
    \caption{Framework overview: Visual observations \( o_s \) and \( o_g \) are encoded to \( v_s \) and \( v_g \). A Task-Classifier predicts a task class and its representation \( \hat{c} \) using \( v_s \) and \( v_g \). These, along with a learnable \(\texttt{START}\) token, are fed into the auto-regressive base planner \( F_{base} \) for sequential action embedding prediction. For each prediction at position \( t \), the retrieval component uses the predicted context vector \( \varsigma_t \) to estimate a probability distribution \( P_{\text{retr,t}} \) over the closest \( K \) key-value pairs in memory for kNN linear interpolation with \( P_{\text{base,t}} \) to estimate the final probability distribution \( P_{\text{all,t}} \).}
    \label{fig:model}
\end{figure*}

\begin{figure*}[t]
    \centering
    \includegraphics[width=0.7\linewidth]{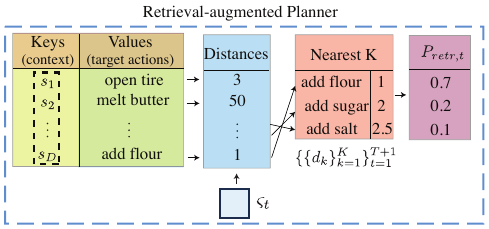}
    \caption{The Retrieval-Augmented Planner module uses the predicted context vector \( \varsigma_t \) to retrieve the \( K \) closest keys and their values from memory, estimating a probability distribution \( P_{\text{retr,t}} \) over these entries for next action probability estimation \( P_{\text{all,t}} \).}
        \label{fig:model_retrv}
\end{figure*}

We propose a new framework, Retrieval-Augmented Planner (RAP), to tackle the adaptive procedure planning problem.
\cref{fig:model} illustrates our framework. 
As the first step, we obtain state $v_i$, step $l_j$, and task $c$ representations from visual observation $o_i$, action step $a_j$, and task title $tsk$ (\eg "make pancakes") respectively. This is done by inputting $o_i$, $a_j$, and $tsk$ to a pretrained vision and language encoder respectively (\cite{miech2020end}), similar to P3IV~\cite{p3iv}.\\

\noindent\textbf{Task Classifier}. Recent studies \cite{wang2023pdpp, wang2023event} have demonstrated the effectiveness of integrating task information into training to enhance model performance in procedure planning. In alignment with previous work \cite{wang2023pdpp}, we adopt a similar strategy by first training an MLP task classifier. This classifier takes initial and final visual observations as inputs and predicts the task class $\hat{tsk}$. The training is conducted using a cross-entropy loss function with the ground-truth task class $tsk$. Subsequently, we incorporate the task representation $\hat{c}$ corresponding to the predicted task class $\hat {tsk}$ as an additional input to both the Base-Planner and the Retrieval-Planner.\\

\noindent\textbf{Base Planner}. To facilitate the adaptive procedure planning setting, we utilize an auto-regressive transformer \cite{radford2019language} as the base planner $F_\text{base}$. 
The base planner takes the representations of the initial state $v_s$, the goal state $v_g$, the predicted task $\hat c$, and the current action $a_{t}$, as input, to predict the next action $a_{t+1}$, \ie, $P_\text{base}(y|x)=P(a_{t+1}|a_t,v_s,v_g, \hat c)$.
The sequence starts with $a_0=\texttt{START}$ and completes upon predicting $a_{T+1}=\texttt{END}$.
Its implementation includes a learnable positional embedding $p_t$ for sequence generation. Therefore, The input sequence to predict $t$-th action step $a_t$ is formulated as in \cref{eq:positional_embed},
\begin{equation}
  [v_s, v_g, \hat c, l_{\texttt{START}}, l_{1} + p_1, \dots,l_{t-2} + p_{t-2}, l_{t-1}+p_{t-1}]
  \label{eq:positional_embed}
\end{equation}
and the output sequence for supervision is~\cref{eq:positional_embed_output},
\begin{equation}
  l=[l_1, l_2, \dots,l_T, l_{\texttt{END}}]\label{eq:positional_embed_output}
\end{equation}
where $l_t$ represents the language representation for step $a_t$, $l_{\texttt{START}}$ and $l_{\texttt{END}}$ are the embeddings for \texttt{START} and \texttt{END} tokens. 
The predicted action step $\hat{a}_t$ is determined by mapping to a probability distribution over the action vocabulary $V_a$ (which includes the \texttt{END} token), which is achieved by comparing the language similarity between predicted $\hat{l}_t$ and those within $V_a$, enabling us to assign probabilities to each action within a plan sequence. During the training stage, we use the ground-truth action representation $l_{t}$ as input, while we use the predicted step representation $\hat{l}_{t}$ during the test stage.\\

\noindent\textbf{Retrieval Planner}. 
To improve long-range sequence modeling, our model integrates a direct retrieval mechanism based on a training datastore, utilizing its state-action temporal dependency information, explicitly. This enhancement is achieved by augmenting the base planner in a manner similar to \cite{khandelwal2019generalization, grave2017unbounded}, by adding a retrieval-augmented  memory module, which learns the state-action pairs in the training data as memory, by modeling states, as a context for the actions. \cref{fig:model_retrv} illustrated the workings of this component.

The memory module consists of key-value pairs $\{s_n:d_n\}_{n=1}^{D}$, where $D$ is the memory size. The keys ($s_n$), are discrete learnable context vectors and the values ($d_n$) are their subsequent action step labels, conditioned on the context. We estimate a context vector ${\varsigma_t}$, given representations of the initial visual state $v'_s$, the target state $v'_g$, the current action $l_t$, and a learnable positional embedding $q_t$ as in~\cref{eq:plan_state}
\begin{equation}
  \varsigma_t = f_{\text{Proj}}( \text{Concat}(v_s, v_g, \hat c, l_t+q_t))
  \label{eq:plan_state}
\end{equation}
With $f_{\text{Proj}}$ as a projection layer and $\text{Concat}(\cdot)$ concatenating inputs along the embedding dimension. In training, we learn the keys ($s_n$) by comparing them to inputs' context vectors $\varsigma_t$. We construct the memory values ($d_n$) by uniformly sampling actions from the training dataset (including the \texttt{END} token), making sure of a balanced representation of actions. At inference, the model retrieves the $K$ nearest (\ie, most similar) keys $\{s_k\}_{k=1}^K$ from the memory to a given current context $\varsigma_t$, returning their corresponding values $\{d_k\}_{k=1}^K$. 
The nearest neighbors ($\{s_k\}_{k=1}^K$) get selected by calculating the cosine distance between $\varsigma_t$ and the keys of the datastore $\{s_n\}_{n=1}^D$.
Upon identifying the nearest neighbors, a probability distribution across the retrieved values ($\{d_k\}_{k=1}^K$) can be calculated by applying a softmax function to the similarity scores. 
The retrieval process can be formulated as $P_{\text{retr}}(y|x)=P(a_{t+1}|\varsigma_t,\ a_{t+1} \in \{d_k\}_{k=1}^K)$.
The final prediction ensembles the predictions from base planner and retrieval planner. Following \cite{grave2017unbounded, khandelwal2019generalization}, we obtain the final probability distribution for the next action step as:
\begin{equation}
  P_{\text{all}}(y|x) = \lambda P_{\text{retr}}(y|x) + (1 - \lambda) P_{\text{base}}(y|x)
  \label{eq:prob_aug}
\end{equation}
where $\lambda$ is a tuned hyper-parameter.
\subsection{Training} \label{ssec: training}

RAP is trained in two stages: First, we train the base planner, then we augment and train the model including the retrieval planner. This approach ensures effective learning of the retrieval planner's keys, building upon the base planner's good initialization obtained from the initial training stage.\\
\noindent {\bf Stage 1: Training the Base Planner}. During the training phase, we first supervise the base model's output, the textual representation of each action step in the procedure sequence, through language supervision combined with contrastive learning \cite{gutmann2010noise}: For each feature $\hat{l}_t$ generated by the transformer's head, we take its ground truth language embedding $l_t$ as positive sample and all other embeddings in the action vocabulary $V_a$ as negative examples. The loss is formulated as:
\begin{equation}
  \mathcal{L}_{\text{base}} = - \sum_{t=1}^{T+1} \log \frac{\exp{\hat{l}_t \cdot l_t}}{\sum_j \exp{\hat{l}_t \cdot l_j}}
  \label{eq:contra_loss}
\end{equation}
Where ($T$+1)-th step denotes the \texttt{END} token.\\

\noindent {\bf Stage 2: Training the Base and Retrieval Planners}. 
In the second stage, the model is augmented with the nearest neighbor retrieval-based memory. The retrieval planner advances learning by juxtaposing the one-hot ground-truth labels $a_t \in p$, with the model’s interpolated probability estimates $P_{all,t}$, for each action at each time-step $t$. It refines the training by assessing how likely each action is, based on the model’s predictions, instead of solely relying on matching the correct action. We achieve this, using a cross-entropy minimization in \cref{eq:mem_loss},
\begin{equation}
  \mathcal{L}_{\text{all}} = - \sum_{t=1}^{T+1} a_t \log P_{\text{all,t}}(y|x)
  \label{eq:mem_loss}
\end{equation}
where $P_\text{all,t}(y|x)$ is defined based on Eq.~\eqref{eq:prob_aug}.

\subsection{Addition of Unannotated Videos with Weak Supervision}\label{sec:weak_sup}
A significant challenge in procedure planning for instructional videos is the scarcity of annotated data. This is mainly because annotating instructional videos with step-level (\ie, timestamps) or sequence-level labels (\ie, action sequences) is resource-intensive. To address this issue, we propose a weakly-supervised training approach aimed at expanding the annotated training data to include additional unannotated task-related videos in the training process.  

Specifically, we align instructional videos with their corresponding topics, such as ``making an omelet'' or ``making pancakes''. 
This alignment links each video subset to a subset of instructional action plans sharing a similar topic. Crucially, this approach does not require detailed annotations that map specific action steps to individual video segments or frames. Instead, it relies on the broader thematic association between the video content and the instructional plans, allowing for a more general form of supervision during training. However, the acquired pairs of instructional videos and action plans may exhibit certain discrepancies. First, the presence of action steps in a plan that do not appear in the related video. Second, a mismatch in the sequence of action steps between the plan and their occurrence in the instructional video. Third, the inclusion of additional action steps in the video that are absent from the plan.

To tackle these challenges, we applied a video grounding algorithm, taking inspiration from Dvornik \etal~\cite{dvornik2021drop} to generate pseudo-annotations of descriptions in instructional videos. 
In short, our grounding algorithm processes an instructional video alongside an ordered sequence of action steps to produce a sequence of grounding results ${(t_{start}, t_{end}, a)}$, where $t_{start}$ and $t_{end}$ denote the beginning and end times of action step $a$. The grounding model offers flexible sequence alignment and the ability to exclude irrelevant actions or frames, effectively addressing observed discrepancies. Further details on the grounding model are provided in the appendix.
We utilize these pseudo-annotations, extracted for each relevant video, which serve as structurally and format-wise equivalent to manually annotated videos. These pseudo-annotations are integrated into Stage 1 training alongside original annotations, where the model is tasked with predicting the complete sequence of actions based on the starting frame of the first action and the ending frame of the last action outlined in the plan.



This weak supervision of unannotated data, is an extension of Stage 1 training by using both original annotations of training videos and pseudo-annotations of task-relevant videos as the new training set. This training approach is a step towards a cost-effective solution for generalizing and scaling procedure planning to abundant unannotated videos.

\section{Experiments}
\label{sec:Experiments}
We experiment on two benchmark datasets and employ four metrics to validate the efficacy of our proposed model.
\begin{figure}{}
    \centering
    \includegraphics[width=0.8\textwidth]{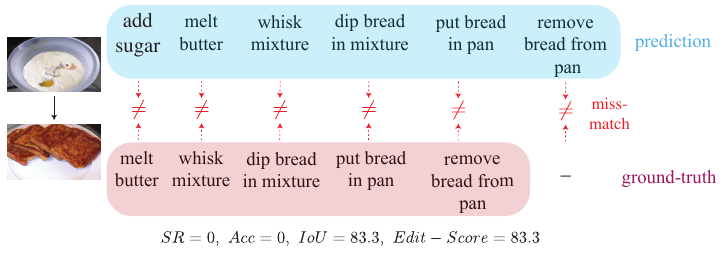}
    \caption{Illustration of the advantage of edit score, in evaluation of long varied-length sequences. As it can be seen, despite the prediction being plausible and acceptable, accuracy and success rate, falsely flag this scenario as a miss.}
    \label{fig:metric_fail}
\end{figure}
\subsection{Datasets}
Our methodology is evaluated using two instructional video datasets: CrossTask \cite{crosstask} and COIN \cite{coin}. CrossTask contains 2750 videos spanning 83 different topics, such as "How to make french toast?", with 7.6 actions per video on average. These topics are categorized into 18 primary and 65 related events. The primary topics are detailed with precise timestamps for each action step, providing a clear sequence of instructional activities. In contrast, the related instructional topics, gathered through automatic collection methods, lack annotations for action-step timestamps, sequence order, and the specific subset of action steps each video covers. Each topic within this related set is associated with a designated instructional plan as well as a set of unannotated instructional videos of the same topic. These videos may exhibit variations from the outlined plan, as detailed in \cref{ssec: training}, due to the previously mentioned discrepancies. We use CrossTask's related-set data, during weakly-supervised training with language supervision. COIN, includes 11827 videos, covering 778 different actions, with an average action-horizon of 3.6 actions per video. We follow \cite{chang2020, p3iv} by creating train/test splits with a 70\%/30\% ratio as well as implementing a moving window approach to organize the datasets into plans with various action-horizons.

\subsection{Metrics}
\label{sec:metrics}

Our methodology's performance is evaluated using three standard metrics \cite{chang2020, p3iv, sun2022plate, bi2021}. First, mean Intersection over Union ($mIoU$) assesses overlap between predicted and ground truth action sequences, calculated by $\frac{|{a_t}\cap{\hat{a}_t}|}{|{a_t} \cup {\hat{a}_t}|}$. This metric evaluates whether the model identifies correct steps but does not consider action order or repetitions. Second, mean Accuracy ($mAcc$) measures action alignment at each step, balancing the impact of repeated actions. Lastly, the Success Rate ($SR$) evaluates a plan as successful if it exactly matches the ground truth.

However, Accuracy and Success Rate have limitations, particularly with long or variable-length sequences, often leading to false negatives due to their sensitivity to action order. For example, in a pancake-making plan, if the predicted sequence adds a plausible step like ``Add sugar'' at the beginning, accuracy may incorrectly deem it inaccurate. This issue is more pronounced with sequences that vary in length from the ground truth, indicating the need for a more nuanced evaluation metric in procedure planning that accommodates plausible variations in actions.

To address the limitations of the current metrics, we introduce a new metric, \textit{mean edit-score} ($mES$), based on the minimum edit distance (\eg Levenshtein edit distance), measuring the similarity between two sequences. This function, $L(s_1, s_2)|{c_{sub}, c_{ins}, c_{del}}$, calculates the minimum number of substitutions ($c_{sub}$), insertions ($c_{ins}$), and deletions ($c_{del}$) needed to transform sequence $s_1$ into $s_2$. The mean edit-score, is derived from the expected value of sample edit-scores. The edit-score is thus defined based on the optimal edit distance as:

\begin{equation}
 ES_{s_1, s_2} = 100 \cdot \frac{\max{(\ |s_1|,\ |s_2|\ )}-L(s_1, s_2)|_{1, 1, 1}}{\max{(\ |s_1|,\ |s_2|\ )}},
\label{eq:edt_c}
\end{equation}
Where the operator $|\cdot |$ measures the length of a sequence. 
The mean edit-score offers a more flexible and forgiving assessment compared to strict metrics like success rate and mean accuracy, while still maintaining sensitivity to the order of action steps, hence complimenting $mIoU$ (\ie, if the ground-truth is \texttt{add curry leaves, stir mixture, add chili powder} and the prediction is \texttt{add chili powder, stir mixture, add curry leaves}, we face a high $IoU=100\%$ and a low $ES=33.33\%$). This metric provides a more comprehensive evaluation, especially in scenarios where the predicted sequence is plausible yet varies in actions or length from the ground truth (\cref{fig:metric_fail}).

\subsection{Implementation Details} \label{sec:main_implement}
Consistent with earlier approaches, we encode the start and goal visual observations using the S3D network \cite{miech2020end} pre-trained on the
HowTo100M \cite{miech2019howto100m} dataset. For optimization, we use the AdamW optimizer \cite{loshchilov2017decoupled}, setting the weight decay to 0.2. The learning rate is initially set at $lr=0.001$, and we apply a reduce-on-plateau scheduler to adjust this rate based on performance metrics. We experiment with an autoregressive transformer with 7 layers and 8 heads. 
We choose weight $\lambda=0.1$ and volume $D=1000$, for the memory module. Further implementation details can be found in the supplementary section.

\subsection{Comparison with State-of-The-Art Baselines}
\begin{table}[t]
\centering
\caption{Assessment of procedure planning on CrossTask for action-horizons \( T=3 \) and \( T=4 \). Vision supervision is denoted by V, language supervision by L, and weakly-supervised training by +wL.}
\label{tab:crosstask_soa}
\resizebox{0.9\textwidth}{!}{%
\begin{tabular}{cccccccc}
\hline
\multirow{2}{*}{Models}            & \multirow{2}{*}{Supervision} & \multicolumn{3}{c}{T=3}                           & \multicolumn{3}{c}{T=4}                           \\ \cline{4-4} \cline{7-7}
                                   &                              & $SR \uparrow$ & $mAcc \uparrow$ & $mIoU \uparrow$ & $SR \uparrow$ & $mAcc \uparrow$ & $mIoU \uparrow$ \\ \hline
Random                             & -                            & $<$0.01       & 0.94            & 1.66            & $<$0.01       & 0.83            & 1.66            \\
Retrieval-Based                    & V                            & 8.05          & 23.30           & 32.06           & 3.95          & 22.22           & 36.97           \\
WLTDO\cite{ehsani2018let}          & V                            & 1.87          & 21.64           & 31.70           & 0.77          & 17.92           & 26.43           \\
UAAA\cite{abu2019uncertainty}      & V                            & 2.15          & 20.21           & 30.87           & 0.98          & 19.86           & 27.09           \\
UPN\cite{srinivas2018universal}    & V                            & 2.89          & 24.39           & 31.56           & 1.19          & 21.59           & 27.85           \\
DDN\cite{chang2020}                & V                            & 12.18         & 31.29           & 47.48           & 5.97          & 27.10           & 48.46           \\
Ext-GAIL w/o Aug.\cite{bi2021}     & V                            & 18.01         & 43.86           & 57.16           & -             & -               & -               \\
Ext-GAIL\cite{bi2021}              & V                            & 21.27         & 49.46           & 61.70           & 16.41         & 43.05           & 60.93           \\ \hline
P3IV w/o Adv.\cite{p3iv}           & L                            & 22.12         & 45.57           & 67.40           & -             & -               & -               \\
P3IV \cite{p3iv}                   & L                            & 23.34         & 49.96           & 73.89           & 13.40         & 44.16           & 70.01           \\
E3P (baseline)\cite{wang2023event} & L                            & 22.56         & 46.17           & -               & 12.97         & 43.97           & -               \\
E3P \cite{wang2023event}           & L                            & 26.40         & 53.02           & 74.05           & 16.49         & 48.00           & 70.16           \\
Skip-Plan \cite{li2023skip}        & L                            & 28.85         & 61.18           & 74.98           & 15.56         & 55.64           & 70.30           \\
\textbf{RAP} {\small(ours w/o unannotated data)} &
  L &
  \textbf{29.23} &
  \textbf{62.35} &
  \textbf{75.63} &
  \textbf{16.96} &
  \textbf{55.77} &
  \textbf{71.49} \\
\textbf{RAP} {\small(ours with unannotated data)} &
  L+wL &
  \textbf{30.51} &
  \textbf{62.68} &
  \textbf{75.92} &
  \textbf{17.10} &
  \textbf{55.91} &
  71.33 \\ \hline
\end{tabular}%
}
\end{table}

\begin{table}[]
\centering
\caption{Comparison of RAP with PDPP, when we align our state definition with PDPP's for CrossTask and \( T=3 \).}
\label{tab:RAP_PDPP}
\resizebox{0.75\textwidth}{!}{%
\begin{tabular}{ccccc}
Models                                                       & Supervision & $SR \uparrow$  & $mAcc \uparrow$ & $mIoU \uparrow$ \\ \hline
PDPP \cite{wang2023pdpp} {$\clubsuit$}                       & L           & 37.20          & 64.67           & 66.57           \\
\textbf{RAP} {\small(ours w/o unannotated data) $\clubsuit$} & L           & \textbf{38.51} & \textbf{65.18}  & \textbf{77.44}  \\ \hline
\end{tabular}%
}
\end{table}

\begin{table}[t]
\centering
\caption{Comparison of the procedure planing models' performance on the COIN dataset.}
\label{tab:coin_soa}
\resizebox{0.75\textwidth}{!}{%
\begin{tabular}{ccccccc}
\hline
\multirow{2}{*}{Models}             & \multicolumn{3}{c}{T=3}                            & \multicolumn{3}{c}{T=4}                            \\ \cline{3-3} \cline{6-6}
                                    & $SR \uparrow$  & $mAcc \uparrow$ & $mIoU \uparrow$ & $SR \uparrow$  & $mAcc \uparrow$ & $mIoU \uparrow$ \\ \hline
Random (V)          & $<$0.01 & $<$0.01 & 2.47  & $<$0.01 & $<$0.01 & 2.32  \\
Retrieval-Based (V) & 4.38    & 17.40   & 32.06 & 2.71    & 14.29   & 36.97 \\
DDN (V)             & 13.90   & 20.19   & 64.78 & 11.13   & 17.71   & 68.06 \\
P3IV (L)            & 15.40   & 21.67   & 76.31 & 11.32   & 18.85   & 70.53 \\
E3P (L)             & 19.57   & 31.42   & 84.95 & 13.59   & 26.72   & 84.72 \\
PDPP (L)            &
21.33   &
45.62   &
51.82 &
14.41   &
44.10   &
51.39 \\
Skip-plan (L)       & 23.65   & 47.12   & 78.44 & 16.04   & 43.19   & 77.07 \\
\textbf{RAP} ( w/o unannotated) (L) & \textbf{24.47} & \textbf{47.59 } & \textbf{85.24}  & \textbf{16.76} & 42.11           & 84.64   \\ \hline

\end{tabular}%

}
\end{table}

\begin{table*}[h]
\centering
\caption{Evaluation of RAP versus a retrieval-based (RB) model in variable-length sequence prediction on CrossTask and COIN.}
\resizebox{0.9\textwidth}{!}{%
\begin{tabular}{ccccccccc}
\hline  
                            & \multicolumn{4}{c}{CrossTask}                         & \multicolumn{4}{c}{COIN}         \\ \cline{2-9} 
\multirow{-2}{*}{Models} &
  $SR \uparrow$ &
  $mAcc \uparrow$ &
  $mIoU \uparrow$ &
  $mES \uparrow$ &
  $SR \uparrow$ &
  $mAcc \uparrow$ &
  $mIoU \uparrow$ &
  $mES \uparrow$ \\ \hline
\multicolumn{1}{c|}{Random} & 0.02 & 0.56  & 1.52  & \multicolumn{1}{c|}{1.21}  & 0.02 & 0.09  & 0.18  & 0.17  \\
\multicolumn{1}{c|}{RB}     & 8.01 & 21.32 & 38.80 & \multicolumn{1}{c|}{33.43} & 7.11 & 12.94 & 20.17 & 17.69 \\
\multicolumn{1}{c|}{\textbf{RAP} (w/o unannotated data)} &
  {\color[HTML]{1A1C1F} \textbf{20.28}} &
  {\color[HTML]{1A1C1F} \textbf{49.41}} &
  \textbf{69.95} &
  \multicolumn{1}{c|}{{\color[HTML]{1A1C1F} \textbf{62.79}}} &
  \textbf{19.14} &
  \textbf{33.26} &
  \textbf{75.82} &
  \textbf{57.48} \\
\multicolumn{1}{c|}{\textbf{RAP} (with unannotated data)} &
  {\color[HTML]{1A1C1F} \textbf{21.79}} &
  {\color[HTML]{1A1C1F} \textbf{50.58}} &
  \textbf{71.44} &
  \multicolumn{1}{c|}{{\color[HTML]{1A1C1F} \textbf{63.86}}} &
  - &
  - &
  - &
  - \\
\hline  
\end{tabular}%
}
    \label{tab:var_mem}
\end{table*}

Our study conducts a comprehensive comparison of our new framework, RAP, against established models in procedure planning, using CrossTask and COIN datasets. We refer to our framework without the the retrieval-based memory augmentation, \ie, base planner, as BP (as opposed to RAP). We categorize previous approaches based on their supervision: language (L) or vision (V). Our proposed model's performance compared to the previous works, is detailed in \cref{tab:crosstask_soa,tab:RAP_PDPP} for CrossTask, and \cref{tab:coin_soa} for COIN.

The inherent uncertainty in variable action-horizon procedure planning models complicates direct and fair comparisons with traditional fixed-length models, due to the added complexity of predicting the conclusion of a sequence as well as the broader range of acceptable sequences of variable length other than the ground-truth. To ensure a fair evaluation, we compare our variable action-horizon architecture, trained and evaluated exclusively on fixed-horizon data (of length $T$), to the fixed-horizon models (of length $T$) from prior studies in procedure planning \cite{ehsani2018let,abu2019uncertainty,  srinivas2018universal, chang2020, bi2021,p3iv, wang2023event, wang2023pdpp, li2023skip}. \cref{tab:crosstask_soa} further shows the performance boost achieved by the addition of CrossTask's unannotated data (\ie, CrossTask's related video set) through weakly-supervised training (supervision: L + wL). 

Our findings demonstrate that RAP exhibits superior performance over existing vision-supervised models. 
Notably, RAP, without the addition of unannotated data, surpasses Ext-GAIL \cite{bi2021} with a 7.97\% increase in Success Rate (SR), 12.89\% in Mean Accuracy (mAcc), and 13.93\% in Mean Intersection over Union (mIoU) for sequence lengths of T=3 in CrossTask.

When compared to language-supervised models like P3IV, E3P, and Skip-Plan, RAP demonstrates superior performance on CrossTask. For T=3, RAP exhibits improvements of 5.99\%, 2.83\%, and 0.38\% respectively, and for T=4, it shows enhancements of 3.56\%, 0.47\%, and 1.4\% in SR. \cref{tab:RAP_PDPP} compares our model with PDPP \cite{wang2023pdpp}, which employs a different procedure state definition. For fairness, we evaluate our model under the PDPP setting, marked with $\clubsuit$, showcasing its superior performance. PDPP uses a 3-second window following the onset and preceding the offset of action steps to capture visual states, giving it an advantage in accessing action-related information compared to previous models. Traditionally, a 1-second window centered around the start and end of actions is used. By aligning our state definition with PDPP's, our RAP model outperforms PDPP on all metrics, as shown in the table. We observe a similar superiority in performance, when evaluating the model on the COIN dataset as shown in \cref{tab:coin_soa}.

\begin{table}[t]
\centering
\caption{A study on the effect of retrieval augmentation (BP vs RAP), for variable and fixed action-horizon settings.}
\label{tab:crosstask_mem}
\resizebox{1\textwidth}{!}{%
\begin{tabular}{|c|c|cccc|cccc|cccc|}
\hline
\multirow{2}{*}{Dataset} &
  \multirow{2}{*}{Models} &
  \multicolumn{4}{c|}{T=3} &
  \multicolumn{4}{c|}{T=4} &
  \multicolumn{4}{c|}{Variable Length T} \\ \cline{4-5} \cline{8-9} \cline{12-13}
 &
   &
  $SR \uparrow$ &
  $mAcc \uparrow$ &
  $mIoU \uparrow$ &
  $mES \uparrow$ &
  $SR \uparrow$ &
  $mAcc \uparrow$ &
  $mIoU \uparrow$ &
  $mES \uparrow$ &
  $SR \uparrow$ &
  $mAcc \uparrow$ &
  $mIoU \uparrow$ &
  $mES \uparrow$ \\ \hline
\multirow{2}{*}{\begin{tabular}[c]{@{}c@{}}CrossTask\\ (w/o unannotated)\end{tabular}} &
  BP &
  25.42 &
  58.70 &
  73.91 &
  65.59 &
  15.23 &
  52.16 &
  70.53 &
  61.63 &
  18.54 &
  47.85 &
  69.16 &
  62.64 \\
 &
  RAP &
  \textbf{29.23} &
  \textbf{62.35} &
  \textbf{75.63} &
  \textbf{67.33} &
  \textbf{16.96} &
  \textbf{55.77} &
  \textbf{71.49} &
  \textbf{64.51} &
  \textbf{20.28} &
  \textbf{49.41} &
  \textbf{69.95} &
  \textbf{62.79} \\ \hline
\multirow{2}{*}{COIN} &
  BP &
  23.38 &
  46.21 &
  80.49 &
  62.28 &
  14.29 &
  41.67 &
  78.62 &
  57.26 &
  17.86 &
  31.34 &
  72.44 &
  56.24 \\
 &
  RAP &
  \textbf{24.47} &
  \textbf{47.59} &
  \textbf{85.24} &
  \textbf{63.15} &
  \textbf{16.76} &
  \textbf{42.11} &
  \textbf{84.64} &
  \textbf{60.47} &
  \textbf{19.14} &
  \textbf{33.26} &
  \textbf{75.82} &
  \textbf{57.48} \\ \hline
\end{tabular}%
}
\end{table}

\begin{figure}[t]
    \centering
    \includegraphics[width=0.9\linewidth]{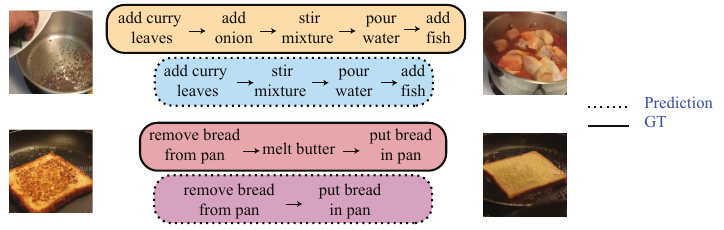}
    \caption{Sample plausible predicted plan sequences generated by RAP vs the ground-truth (GT). }
    \label{fig:sample_res}
\end{figure}

We assessed the efficacy of RAP in handling tasks with variable lengths by conducting experiments against a retrieval-based model, which operates under a similar variable action-horizon setting. The results of this comparison are presented in \cref{tab:var_mem}.
As demonstrated in the table, RAP proves a remarkable superiority over the retrieval-based model across all evaluated metrics. This is particularly evident in the mES metric, where RAP (w/o unannotated data) shows a significant improvement by 29.36\%. 

On CrossTask, RAP is further trained using the weak language supervision framework outlined in \cref{ssec: training}, utilizing the unannotated data from the dataset's related-tasks. As shown in \cref{tab:crosstask_soa} and  \cref{tab:var_mem} when training RAP with the addition of the unannotated data, we see significant improvement in performance across all metrics.

\cref{fig:sample_res} displays two qualitative results of RAP's performance in predicting plans of varying lengths compared to the ground truths. In the top row, it accurately identifies the addition of fish, other ingredients, and the plausible action "add onions". In the bottom row, it recognizes the completed bread, its removal, and suggests the reasonable action "melt butter", before the new bread is put in the pan, showcasing the model's planning abilities.

\subsection{Ablation Study}

{ \bf Impact of retrieval augmentation}. The introduction of a retrieval-based memory component significantly enhances our framework's performance. \cref{tab:crosstask_mem} demonstrates improvements for both fixed and variable action-horizon settings. Notably, enhancements include increases of 3.81\%, 1.73\%, and 1.74\% in Success Rate (SR) for T=3 and T=4 in the fixed setting, and in the variable setting, respectively, on CrossTask. We observe a similar boost in performance for the COIN dataset, as evidenced in the \cref{tab:crosstask_mem}. \cref{tab:crosstask_mem_T} details the enhancements in SR, categorized by the ground-truth (GT) sequences' action-horizon in the variable-length sequence prediction setting.
\begin{table}[t]
\centering
\caption{Impact of retrieval augmentation (BP vs RAP) in $SR$, on CrossTask, categorized by ground-truth (GT) action-horizon in the variable action-horizon setting.}
\label{tab:crosstask_mem_T}
\resizebox{0.65\textwidth}{!}{%
\begin{tabular}{ccccccc}
\hline
\multirow{2}{*}{\begin{tabular}[c]{@{}c@{}}Model\\ (w/o unannotated data)\end{tabular}} &
  \multicolumn{6}{c}{GT action-horizon} \\ \cline{2-7} 
   & 1     & 2     & 3     & 4    & 5    & 6    \\ \hline
BP & 49.58 & 27.84 & 16.21 & 8.37 & 5.89 & 3.34 \\
RAP &
  \textbf{52.60} &
  \textbf{29.20} &
  \textbf{18.02} &
  \textbf{10.52} &
  \textbf{6.16} &
  \textbf{5.18} \\ \hline
\end{tabular}%
}
\end{table}
\begin{table}[t]
\centering
\caption{Weak language supervision impact: Table shows RAP's performance gains on CrossTask with increasing unannotated data use.}
\label{tab:pretrain_eff}
\resizebox{0.6\textwidth}{!}{%
\begin{tabular}{ccccc}
\hline
\begin{tabular}[c]{@{}c@{}}Used unannotated\\data portion\end{tabular} & $SR \uparrow$ & $mAcc \uparrow$ & $mIoU \uparrow$ & $mES \uparrow$ \\ \hline
No unannotated data & 20.28          & 49.41          & 69.95          & 62.79          \\
25\%                & 19.20          & 47.88          & 69.05          & 61.96          \\
50\%                & 20.57          & 49.72          & 69.75          & 63.12          \\
75\%                & 21.31          & 50.22          & 70.86          & 63.40          \\
100\%               & \textbf{21.79} & \textbf{50.58} & \textbf{71.44} & \textbf{63.86} \\ \hline
\end{tabular}%
}
\end{table}
\\
{ \bf Impact of weak language supervision}. The influence of weak language supervision on our model's performance is evaluated through an ablation study on CrossTask, where we incrementally increased the unannotated data used in weakly-supervised training. \cref{tab:pretrain_eff} displays the varying percentages of unannotated data utilized. The results show consistent performance gains for RAP under weak language supervision, particularly as more data is introduced (except at the 25\% data level). This suggests potential for further enhancements with increased unannotated data. Notably, using 100\% of unannotated data yields improvements in SR, mAcc, mIoU, and mES by 1.51\%, 1.17\%, 1.49\%, and 1.07\%, respectively.

\section{Conclusion}
\label{sec:conclusion}

In this study, we addressed the challenge of procedure planning in instructional videos by developing the Retrieval-Augmented Planner (RAP) model. This model overcomes the limitations of previous approaches that were restricted to fixed action-step predictions. RAP excels in both fixed and varied action horizon scenarios, outperforming existing models. A key innovation is our weak language supervision method, which uses thematic links between unannotated videos and plans for pseudo-annotation generation, enhancing model training. Our model also incorporates a learnable retrieval-based memory component for improved long-range sequence predictions and efficient use of temporal data. Additionally, we introduced ``edit-score'', a new metric for assessing variable-length procedures. Looking forward, we see potential in applying RAP to more diverse scenarios and in generating probabilistic, varied-length plans. We also envisage that expanding our weak supervision approach to include an even wider range of unannotated data could markedly elevate the performance of procedure planning models, paving the way for more advanced applications in this field.

\clearpage  

%
%
\bibliographystyle{splncs04}
\bibliography{main}

\begin{thebibliography}{10}
\providecommand{\url}[1]{\texttt{#1}}
\providecommand{\urlprefix}{URL }
\providecommand{\doi}[1]{https://doi.org/#1}

\bibitem{abu2019uncertainty}
Abu~Farha, Y., Gall, J.: Uncertainty-aware anticipation of activities. In: Proceedings of the IEEE/CVF International Conference on Computer Vision Workshops. pp.~0--0 (2019)

\bibitem{bi2021}
Bi, J., Luo, J., Xu, C.: Procedure planning in instructional videos via contextual modeling and model-based policy learning. In: Proceedings of the IEEE/CVF International Conference on Computer Vision. pp. 15611--15620 (2021)

\bibitem{borgeaud2022improving}
Borgeaud, S., Mensch, A., Hoffmann, J., Cai, T., Rutherford, E., Millican, K., Van Den~Driessche, G.B., Lespiau, J.B., Damoc, B., Clark, A., et~al.: Improving language models by retrieving from trillions of tokens. In: International conference on machine learning. pp. 2206--2240. PMLR (2022)

\bibitem{chang2020}
Chang, C.Y., Huang, D.A., Xu, D., Adeli, E., Fei-Fei, L., Niebles, J.C.: Procedure planning in instructional videos. In: European Conference on Computer Vision. pp. 334--350. Springer (2020)

\bibitem{dvornik2021drop}
Dvornik, M., Hadji, I., Derpanis, K.G., Garg, A., Jepson, A.: Drop-dtw: Aligning common signal between sequences while dropping outliers. Advances in Neural Information Processing Systems  \textbf{34},  13782--13793 (2021)

\bibitem{ehsani2018let}
Ehsani, K., Bagherinezhad, H., Redmon, J., Mottaghi, R., Farhadi, A.: Who let the dogs out? modeling dog behavior from visual data. In: Proceedings of the IEEE Conference on Computer Vision and Pattern Recognition. pp. 4051--4060 (2018)

\bibitem{grave2017unbounded}
Grave, E., Cisse, M.M., Joulin, A.: Unbounded cache model for online language modeling with open vocabulary. Advances in neural information processing systems  \textbf{30} (2017)

\bibitem{gutmann2010noise}
Gutmann, M., Hyv{\"a}rinen, A.: Noise-contrastive estimation: A new estimation principle for unnormalized statistical models. In: Proceedings of the thirteenth international conference on artificial intelligence and statistics. pp. 297--304. JMLR Workshop and Conference Proceedings (2010)

\bibitem{hadji2021representation}
Hadji, I., Derpanis, K.G., Jepson, A.D.: Representation learning via global temporal alignment and cycle-consistency. In: Proceedings of the IEEE/CVF Conference on Computer Vision and Pattern Recognition. pp. 11068--11077 (2021)

\bibitem{khandelwal2019generalization}
Khandelwal, U., Levy, O., Jurafsky, D., Zettlemoyer, L., Lewis, M.: Generalization through memorization: Nearest neighbor language models. arXiv preprint arXiv:1911.00172  (2019)

\bibitem{lei2020mart}
Lei, J., Wang, L., Shen, Y., Yu, D., Berg, T.L., Bansal, M.: Mart: Memory-augmented recurrent transformer for coherent video paragraph captioning. arXiv preprint arXiv:2005.05402  (2020)

\bibitem{li2023skip}
Li, Z., Geng, W., Li, M., Chen, L., Tang, Y., Lu, J., Zhou, J.: Skip-plan: Procedure planning in instructional videos via condensed action space learning. In: Proceedings of the IEEE/CVF International Conference on Computer Vision. pp. 10297--10306 (2023)

\bibitem{loshchilov2017decoupled}
Loshchilov, I., Hutter, F.: Decoupled weight decay regularization. arXiv preprint arXiv:1711.05101  (2017)

\bibitem{miech2020end}
Miech, A., Alayrac, J.B., Smaira, L., Laptev, I., Sivic, J., Zisserman, A.: End-to-end learning of visual representations from uncurated instructional videos. In: Proceedings of the IEEE/CVF Conference on Computer Vision and Pattern Recognition. pp. 9879--9889 (2020)

\bibitem{miech2019howto100m}
Miech, A., Zhukov, D., Alayrac, J.B., Tapaswi, M., Laptev, I., Sivic, J.: Howto100m: Learning a text-video embedding by watching hundred million narrated video clips. In: Proceedings of the IEEE/CVF international conference on computer vision. pp. 2630--2640 (2019)

\bibitem{niu2024schema}
Niu, Y., Guo, W., Chen, L., Lin, X., Chang, S.F.: Schema: State changes matter for procedure planning in instructional videos. arXiv preprint arXiv:2403.01599  (2024)

\bibitem{radford2019language}
Radford, A., Wu, J., Child, R., Luan, D., Amodei, D., Sutskever, I., et~al.: Language models are unsupervised multitask learners. OpenAI blog  \textbf{1}(8), ~9 (2019)

\bibitem{srinivas2018universal}
Srinivas, A., Jabri, A., Abbeel, P., Levine, S., Finn, C.: Universal planning networks: Learning generalizable representations for visuomotor control. In: International Conference on Machine Learning. pp. 4732--4741. PMLR (2018)

\bibitem{sun2022plate}
Sun, J., Huang, D.A., Lu, B., Liu, Y.H., Zhou, B., Garg, A.: Plate: Visually-grounded planning with transformers in procedural tasks. IEEE Robotics and Automation Letters  \textbf{7}(2),  4924--4930 (2022)

\bibitem{coin}
Tang, Y., Ding, D., Rao, Y., Zheng, Y., Zhang, D., Zhao, L., Lu, J., Zhou, J.: Coin: A large-scale dataset for comprehensive instructional video analysis. In: Proceedings of the IEEE/CVF Conference on Computer Vision and Pattern Recognition. pp. 1207--1216 (2019)

\bibitem{vaswani2017attention}
Vaswani, A., Shazeer, N., Parmar, N., Uszkoreit, J., Jones, L., Gomez, A.N., Kaiser, {\L}., Polosukhin, I.: Attention is all you need. Advances in neural information processing systems  \textbf{30} (2017)

\bibitem{wang2023event}
Wang, A.L., Lin, K.Y., Du, J.R., Meng, J., Zheng, W.S.: Event-guided procedure planning from instructional videos with text supervision. In: Proceedings of the IEEE/CVF International Conference on Computer Vision. pp. 13565--13575 (2023)

\bibitem{wang2023pdpp}
Wang, H., Wu, Y., Guo, S., Wang, L.: Pdpp: Projected diffusion for procedure planning in instructional videos. In: Proceedings of the IEEE/CVF Conference on Computer Vision and Pattern Recognition. pp. 14836--14845 (2023)

\bibitem{p3iv}
Zhao, H., Hadji, I., Dvornik, N., Derpanis, K.G., Wildes, R.P., Jepson, A.D.: P3iv: Probabilistic procedure planning from instructional videos with weak supervision. In: Proceedings of the IEEE/CVF Conference on Computer Vision and Pattern Recognition. pp. 2938--2948 (2022)

\bibitem{zhong2022training}
Zhong, Z., Lei, T., Chen, D.: Training language models with memory augmentation (2022)

\bibitem{crosstask}
Zhukov, D., Alayrac, J.B., Cinbis, R.G., Fouhey, D., Laptev, I., Sivic, J.: Cross-task weakly supervised learning from instructional videos. In: Proceedings of the IEEE/CVF Conference on Computer Vision and Pattern Recognition. pp. 3537--3545 (2019)

\end{thebibliography}
\clearpage
\section{Supplementary Materials}
\label{sec:sup}

\appendix

The structure of our supplemental material is outlined as follows: \cref{sec:grounding} elaborates on the grounding algorithm we utilized to align unannotated instructional videos with their parallel instructional procedures. In \cref{sec:implementation_details} we expand on the implementation details followed by further ablation studies in \cref{sec:Further Ablation Studies}. Finally, in \cref{sec:visualization} we show more samples of the model's prediction including failure cases.

\section{Grounding Algorithm}
\label{sec:grounding}

To utilize the unannotated data in our proposed weakly supervised framework, we ground an instructional video 
$V=\{f_1,f_2,\cdots,f_T\}$
to the procedure
$P=\{a_1,a_2,\cdots,a_N\}$, its related action plan sequence, 
through a similarity calculation between each frame 
$f_t$
and each action step 
$a_n$
where 
$T$ denotes the number of frames and $N$ denotes the number of action steps. The match-cost \cite{dvornik2021drop, hadji2021representation}, denoted as $C_{t,n}$ and defined in \cref{eq:plan_vid_sim}, measures the dissimilarity between the visual representations 
$v_t$
and the language representations 
$l_n$:
\begin{equation}
  C_{t,n}=1- \cos{(v_t, l_n)},
  \label{eq:plan_vid_sim}
\end{equation}\\
where the operator $\cos(\cdot)$ captures the cosine similarity between the embeddings.

Note that some video frames are not related to the plan and some action steps are not groundable in the video. For each video-plan pair, we utilize a threshold value, \textit{drop-cost} \cite{dvornik2021drop}, to filter out unrelated frames from the action plan and unrelated action steps from the video frames. If a frame's match-cost with all of the action steps exceeds this threshold, we discard that frame. Similarly, if an action's match-cost with every frame fails to fall below the drop-cost threshold, we exclude that particular action from the grounding algorithm. We determine the drop-cost variable as a specific percentile of values in the match-cost matrix distribution $C= [C_{t,n}]_{T,N}$, as illustrated in \cref{eq:drop_cost}.
 \begin{equation}
\begin{split}
drop-cost = percentile(C\ ,\ perc)\\
\textit{where}~\ perc \in [0, 100],
\end{split}
\label{eq:drop_cost}
\end{equation}
Here, $perc$ is a hyper-parameter of the algorithm (\eg, 15\%) that denotes the percentage of values in $C$ smaller than the corresponding drop-cost.

\begin{figure}{}
    \centering
    \includegraphics[width=0.95\linewidth]{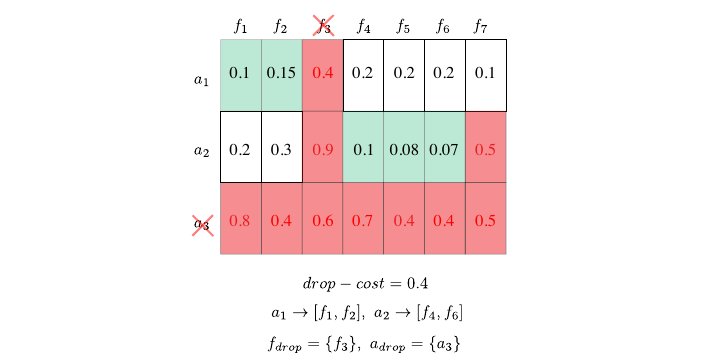}
    \caption{Example illustration on frame to action step alignment. $f_{drop}$ and $a_{drop}$ represent the set of dropped frames and actions. Accordingly, based on the drop-cost value, the cropped sets of frames and actions would be $V'=\{f_1, f_2, f_4,f_6, f_7\}$ and $P'=\{a_1, a_2\}$. In this example, action $a_1$ is matched with frames $[f_1,f_2]$, while action $a_2$ is matched with $[f_4,f_6]$.}
    \label{fig:grounding}
\end{figure}

Let $V'$ and $P'$ denote the sets of frames and action steps after filtering based on the drop-cost, respectively. 
We assign each action step $a'_n \in P'$
with the consecutive frames in $V'$ with the lowest cumulative match-cost sum:

\begin{equation} 
\begin{split}
    start_n,~end_n= &\textit{argmin}\sum_{t=start}^{end} C_{t,n},\\
\textit{such that, }\\
&0 \le start < end \le T,\\
&C_{t,n}<drop-cost\\
\end{split}
\label{eq:frame_assign}
\end{equation}  
where $argmin(\cdot)$ returns the predicted boundaries $(start_n,~end_n)$ such that the cumulative match-cost sum is minimized. \cref{fig:grounding} illustrates an example sequence alignment between frames and action steps, based on the match and drop costs. This grounding algorithm accommodates the three possible discrepancy scenarios mentioned earlier in \cref{sec:weak_sup} effectively.\\

\begin{figure}{}
   \centering
 \includegraphics[width=0.8\columnwidth] {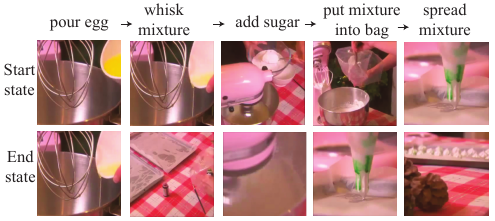}
    \caption{An example of grounding action steps to video to obtain action start and end states.}
   \label{fig:grounding_example}
\end{figure}

\cref{fig:grounding_example} shows an example of grounding instructional videos to the action plans, illustrating the generation of pseudo time-stamps for action start and end points.

\section{Implementation Details}
\label{sec:implementation_details}


To prepare the input data, as detailed in \cref{sec:main_implement}, we employ a pre-trained encoder to extract feature embeddings from action steps and video frames within our plans. These embeddings unify language and visual features in a common space with 512 dimensions. Subsequently, we leverage these embeddings, as discussed in \cref{sec:grounding}, to derive pseudo-annotations for our weakly-supervised training experiments by setting $perc=15\%$ in \cref{eq:drop_cost}.

Following a similar approach to P3IV \cite{p3iv}, we utilize a Multi-Layer Perception (MLP) with dimensions $[512\rightarrow 256 \rightarrow 128]$ to obtain the ground truth language features, denoted as $l_t$. Additionally, we employ another MLP with the same architecture to generate representations of visual states, labeled as $v_t$. To refine our representation of the initial and final visual states, we calculate the average of the representations from three consecutive frames surrounding the ground truth annotations for these states. This involves considering the frame before, the annotated frame serving as the start or end state, and the frame that follows.

Our base planner is based on the 
decoder-only transformer architecture, as described in \cref{sec:main_implement}. It operates with hidden and output dimensions of 128, similar to $v_t$ and $l_t$. Additionally, we employ the $relu(\cdot)$ nonlinearity as the activation function. The memory component, incorporates an MLP layer of $[3*128\rightarrow 256 \rightarrow 128]$ for the projection of the concatenated initial and final visual states ($2*128$) and the current action step representation ($1*128$) according to \cref{eq:plan_state}, and retrives the top $K=10$ items ${\{s_k, d_k\}_{k=1}^K}|_t$ for the $t$th predicted position, as described in \cref{sec:retrieval}. We train RAP in 200 epochs in the the first training stage and 150 epochs in the second training stage with the addition of the retrieval-based memory component. At the test time, the model with best performance on the validation set is used. Section \cref{sec:Further Ablation Studies} includes the ablation study for our design choices.

We used a single \texttt{NVIDIA TITAN RTX} GPU with $24$G of memory for training and evaluation. The inference time is $0.3$s per sample. The training time is $9.1$ hours and $3.8$ hours for stage 1 and stage 2, respectively.

\section{Further Ablation Studies}
\label{sec:Further Ablation Studies}
In this section we present the ablation study that we conducted to determine the the architecture and the hyper-parameters of the model.
\subsection{Ablation of the Base Planner Architecture}

\begin{table}[]
\centering
\caption{Ablation study on the size of the transformer model in the Base Planner on CrossTask.}
\resizebox{0.6\textwidth}{!}{%
\begin{tabular}{cclllll}
\hline
\multirow{4}{*}{Metrics} &
  \multicolumn{6}{c}{\multirow{2}{*}{Number of Layers / Number of Heads}} \\
 &
  \multicolumn{6}{c}{} \\ \cline{2-7} 
 &
  \multicolumn{1}{l}{\multirow{2}{*}{$6 / 4$}} &
  \multirow{2}{*}{$6 / 8$} &
  \multirow{2}{*}{$7 / 4$} &
  \multirow{2}{*}{$7 / 8$} &
  \multirow{2}{*}{$8 / 4$} &
  \multirow{2}{*}{$8 / 8$} \\
 &
  \multicolumn{1}{l}{} &
   &
   &
   &
   &
   \\ \hline
$SR\uparrow$ &
  18.66 &
  \multicolumn{1}{c}{19.25} &
  \multicolumn{1}{c}{19.57} &
  \multicolumn{1}{c}{\textbf{20.28}} &
  \multicolumn{1}{c}{19.74} &
  \multicolumn{1}{c}{19.22} \\
$mES\uparrow$ &
  61.72 &
  \multicolumn{1}{c}{61.89} &
  \multicolumn{1}{c}{62.34} &
  \multicolumn{1}{c}{\textbf{62.79}} &
  \multicolumn{1}{c}{62.42} &
  \multicolumn{1}{c}{62.40} \\ \hline
\end{tabular}%
}  
\label{tab:transformer_size}
\end{table}

We conducted an investigation into the influence of transformer size within the base planner on model performance. As depicted in \cref{tab:transformer_size}, we evaluated the RAP model using variable horizon data derived from CrossTask, and measured its performance based on the success rate and mean edit-score metrics. Our study involved varying the model architecture by adjusting the number of transformer heads and layers. We observed that a model with 7 layers and 8 heads demonstrated the highest level of performance, and the model is robust to the numbers of layers and heads.
\subsection{Ablation of Plan-length Prediction via \texttt{END}}

We study the effectiveness of  the \texttt{END} token in predicting the action-horizon of plans by comparing the plan-length prediction accuracy of BP and RAP against a) random guessing, and b) an MLP classifier serving as a plan-length predictor. As depicted in \cref{tab:tabC}, both RAP and BP significantly outperform the MLP predictor, underscoring the complexity of plan-length prediction. We posit that the \texttt{END} token equips the model with adaptability, ensuring task completion remains contextually relevant without arbitrary constraints, akin to natural human decision-making processes.

For example, consider the scenario of 'cooking pancakes.' Instead of pre-guessing the number of steps (e.g., "five steps to cook pancakes," a task even challenging for humans without listing required actions), our model dynamically adapts to real-time task states. It discerns that after flipping the pancakes, the next logical step might be serving them, rather than rigidly adhering to a predetermined plan-length that fails to accommodate actual cooking progress. This adaptability fosters more accurate and realistic task completion.

\begin{table}[]
\centering
\caption{The accuracy of plan-length prediction for: a) A random model with uniform distribution, b) an MLP length-predictor-module that predicts the length of the plan directly from the initial and final visual inputs, c) the base-planner, and d) RAP.}
\resizebox{0.6\textwidth}{!}{%
\begin{tabular}{ccccc}
\hline
Model          & Random & MLP   & BP    & RAP            \\ \hline
Accuracy  (\%) & 16.25  & 21.32 & 49.77 & \textbf{53.47} \\ \hline
\end{tabular}%
}

\label{tab:tabC}
\end{table}

\subsection{Ablation of the Retrieval Planner Architecture}

\cref{tab:tableA} shows RAP components' performance. The Retrieval head alone under-performs the Base Planner (BP) but can improve BP when combined as RAP. 

To find the optimal architecture and hyper-parameters for the retrieval planner, we conducted an ablation study using the CrossTask dataset. As illustrated in \cref{tab:datastore_size}, we assessed the model's performance in terms of success rate and mean edit-score by varying the size of the Datastore.

Furthermore, we determined the value of the hyper-parameter $\lambda$ in \cref{eq:prob_aug} through experiments conducted on the CrossTask dataset. \cref{tab:lambda_val} details the impact of the $\lambda$ value on model's performance. Overall, the model's performance is not sentitive to the hyper-parameter $\lambda$.

RAP utilizes the retrieval planner in an \textit{explicit} manner by using \textit{context} representation (representing the current task state) to retrieve the most probable \textit{next-actions} from a memory (learnt from the training videos), to enhance base-planner's probability estimations. This is different from prior works' \textit{implicit} incorporation of memory \cite{p3iv}. We compare our explicit memory, to an implicit one of the same size, implemented similarly to P3IV's memory component \cite{p3iv}. ~\cref{tab:A} shows the performance advantage of employing an explicit rather than an implicit memory on CrossTask.

\noindent{\color{teal}\textbf{@R2}} \textbf{Performance of retrieval head $P_{retr}$.} The below table shows RAP components' performance. The Retrieval head alone underperforms the Base Planner (BP) but can improve BP when combined as RAP. 

\begin{table}[]
\centering
\caption{Ablation study on the performance of the retrieval head alone on CrossTask for the setting of variable-length action horizon. }
\label{tab:tableA}
\begin{tabular}{ccccc}
\hline
Model                 & SR (\%)        & mAcc (\%)      & mIoU (\%)      & mES (\%)       \\ \hline
BP $P_{base}$ & 18.54          & 47.85          & 69.16          & 62.64          \\
Retrieval head $P_{retr}$       & 11.42           & 23.28          & 43.67          & 35.96          \\
RAP (w/o unannotated data) & \textbf{20.28} & \textbf{49.41} & \textbf{69.95} & \textbf{62.79} \\ \hline
\end{tabular}
\end{table}

\begin{table}[]
\centering
\caption{Ablation study on the impact of the Datastore size on variable horizon model's performance on CrossTask.}
\begin{tabular}{ccll}
\multirow{4}{*}{Metrics} & \multicolumn{3}{c}{\multirow{2}{*}{Datastore Size}}                                   \\
                         & \multicolumn{3}{c}{}                                                                  \\ \cline{2-4} 
 & \multicolumn{1}{l}{\multirow{2}{*}{500}} & \multirow{2}{*}{1000} & \multirow{2}{*}{1500} \\
                         & \multicolumn{1}{l}{} &                                    &                           \\ \hline
$SR\uparrow$             & 18.82                & \multicolumn{1}{c}{\textbf{20.28}} & \multicolumn{1}{c}{18.03} \\
$mES\uparrow$            & 61.46                & \multicolumn{1}{c}{\textbf{62.79}} & \multicolumn{1}{c}{61.26} \\ \hline
\end{tabular}
\label{tab:datastore_size}
\end{table}

\begin{table}[]
\centering
\caption{Ablation study on the impact of the $\lambda$ value, used for the probability distribution augmentation as described in \cref{eq:prob_aug}, on model's performance on CrossTask.}
\begin{tabular}{ccll}
\multirow{4}{*}{Metrics} & \multicolumn{3}{c}{\multirow{2}{*}{$\lambda$}}                                        \\
                         & \multicolumn{3}{c}{}                                                                  \\ \cline{2-4} 
 & \multicolumn{1}{l}{\multirow{2}{*}{0.05}} & \multirow{2}{*}{0.1} & \multirow{2}{*}{0.15} \\
                         & \multicolumn{1}{l}{} &                                    &                           \\ \hline
$SR\uparrow$             & 19.86                & \multicolumn{1}{c}{\textbf{20.28}} & \multicolumn{1}{c}{19.55} \\
$mES\uparrow$            & 62.32                & \multicolumn{1}{c}{\textbf{62.79}} & \multicolumn{1}{c}{61.94} \\ \hline
\end{tabular}
\label{tab:lambda_val}
\end{table}

\begin{table}[]
\centering
\caption{The abolition study on the addition of RAP vs a Base-Planner with implicit memory of the same size on CrossTask (trained solely on annotated data of fixed action horizon T=3).}
\resizebox{0.6\textwidth}{!}{%
\begin{tabular}{cccc}
Model                      & SR(\%)         & mAcc(\%)       & mIoU(\%)       \\ \hline
BP with Implicit memory    & 26.78          & 60.27          & 74.65          \\
RAP (Results from table 1) & \textbf{29.23} & \textbf{62.35} & \textbf{75.63} \\ \hline
\end{tabular}%
}
\label{tab:A}
\end{table}
\section{Further Visualization}
\label{sec:visualization}

\begin{figure*}
    \centering
    \includegraphics[width=1\linewidth]{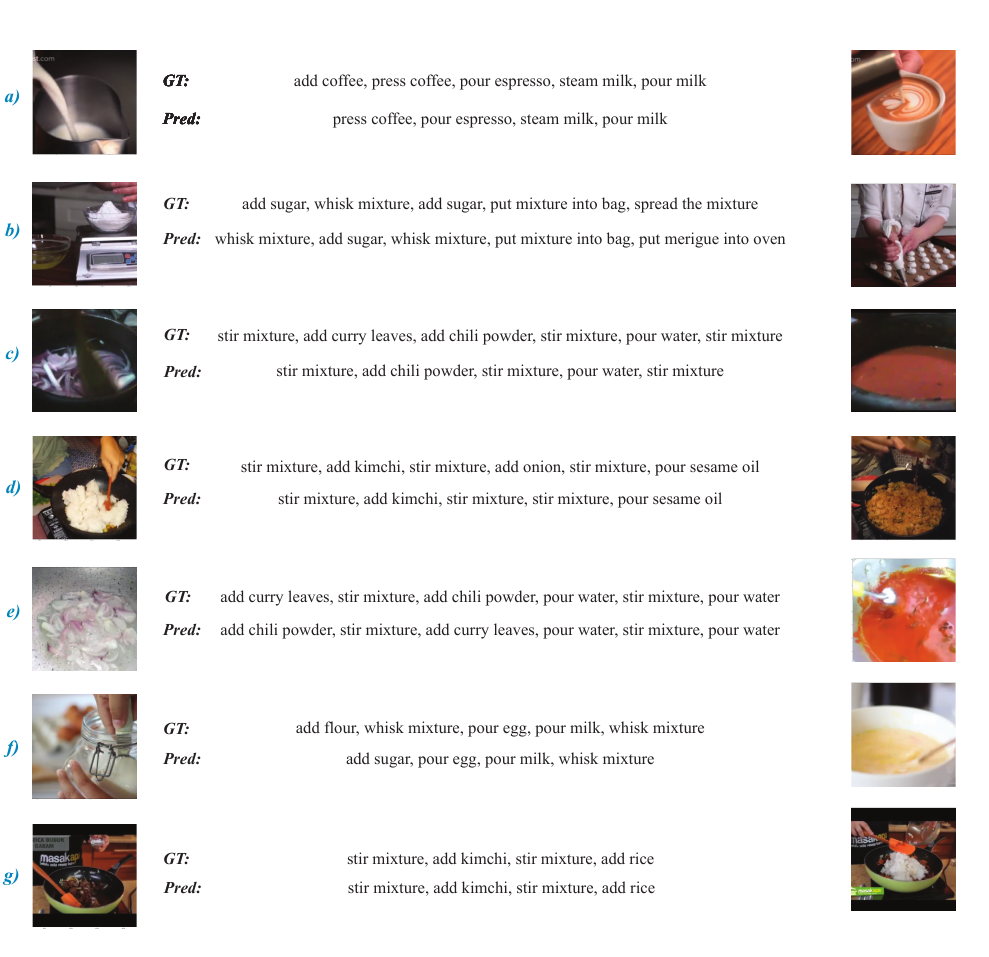}
    \caption{Example predictions of the model trained on CrossTask, based on the initial and target visual observations, where the predicted sequences are of variable lengths.}
    \label{fig:sup_samples}
\end{figure*}

In \cref{fig:sup_samples}, we provide supplementary visual examples of plans generated by RAP, showcasing different prediction horizons. This figure contains some instances where the model encountered challenges and was unable to achieve the ground-truth results. These examples shed light on potential reasons why the model's predictions may diverge from the ground truth:

\begin{itemize}
    \item The model's predictions, while plausible based on the visual initial and target observations, may differ from the ground truth in terms of the number of actions or the specific actions themselves. This can lead to false negatives based on the success-rate metric, as demonstrated in examples \textit{a}, \textit{c}, and \textit{e}.
    
    \item In cases where subtle changes occur in the state of a plan, the model may struggle to understand these nuances. For example, in example \textit{b}, the last step involves spreading a mixture, and the subsequent action should be moving the tray into the oven. However, the model predicts that the mixture spreading is complete and the transfer to the oven has already begun.
    \item The model might overlook certain action steps, especially when their visual impact in the final frame is not clearly discernible. This scenario is evident in example \textit{d}, where the action step "add onion" is absent from the prediction, likely because detecting onions in the final frame poses a visual challenge.
    \item Occasionally, the model may misinterpret the objects present in the visual observations. This is exemplified in example \textit{f}, where the model incorrectly identifies the content of a jar as sugar instead of flour.
\end{itemize}
Finally, example \textit{g}, showcases a scenario where the model predicted the exact sequence as the ground-truth.

\end{document}